# Quantum Cooperative Robotics and Autonomy

Farbod KHOSHNOUD[1,2], Marco B. QUADRELLI[2], Ibrahim I. ESAT[3], and Dario ROBINSON[4]

[1]*Department of Electromechanical Engineering Technology, College of Engineering, California State Polytechnic University, Pomona, CA 91768, USA*

[2]*Mobility and Robotic Systems Section, Jet Propulsion Laboratory, California Institute of Technology, Pasadena, CA, 91109-8099, USA*

[3]*Department of Mechanical and Aerospace Engineering, Brunel University London, Uxbridge UB8 3PH, United Kingdom*

[4]*Police Department, California State Polytechnic University, Pomona, CA 91768, USA*

**Abstract:** The intersection of Quantum Technologies and Robotics Autonomy is explored in the present paper. The two areas are brought together in establishing an interdisciplinary interface that contributes to advancing the field of system autonomy, and pushes the engineering boundaries beyond the existing techniques. The present research adopts the experimental aspects of quantum entanglement and quantum cryptography, and integrates these established quantum capabilities into distributed robotic platforms, to explore the possibility of achieving increased autonomy for the control of multi-agent robotic systems engaged in cooperative tasks. Experimental quantum capabilities are realized by producing single photons (using spontaneous parametric down-conversion process), polarization of photons, detecting vertical and horizontal polarizations, and single photon detecting/counting. Specifically, such quantum aspects are implemented on network of classical agents, i.e., classical aerial and ground robots/unmanned systems. With respect to classical systems for robotic applications, leveraging quantum technology is expected to lead to guaranteed security, very fast control and communication, and unparalleled quantum capabilities such as entanglement and quantum superposition that will enable novel applications.

## 1 Introduction

The research in the area of quantum mechanics in conjunction with robotics applications has been carried out by researchers mainly on developing quantum computers and algorithms, which are able to significantly accelerate the speed of computation (in comparison with classical techniques). There is a vast body of literature indicating the critical importance of quantum computers in advancing various technologies such as robotic applications, although no examples yet exist. One of the key advantages is the envisioned speed of computing. For instance, quantum computing applied to machine intelligence can lead to creating smarter and more creative unmanned systems and robots. The United States NASA uses quantum computing, which is based on the superposition of qubits, for instance, being in superposition of the states of 0 and 1, as well as intermediate states, simultaneously, for communication purposes. This quantum superposition can potentially boost computing power beyond that of any classical computer today. However, the current quantum engineering research and investigations are mainly focused on developing quantum computers and quantum algorithms. Moreover, as quantum computers become accessible in the next decades, for instance for robotic applications, the only rational and practical way for the robots, equipped with quantum computers and quantum computing capabilities, to experience greater performance is to cooperate and communicate with experimental quantum technologies in a multi-robotic network. Experimental quantum communication promises to be the most logical and compatible way for quantum computers networks (e.g., in a network of robotic or unmanned systems) to exchange information. In fact, using classical cooperative robotic techniques between quantum computers, when mounted on robotic platforms in a network, can actually defeat the purpose and advantage of quantum computers and their capability due to the state conversion that is needed to go from the quantum domain to the classical domain and vice versa.

A brief review of the literature in quantum engineering is presented now. The entanglement and superposition capabilities of quantum phenomena are applied in performing quantum computations (e.g., [1], [4]), with quantum bits in superposition of more than one quantum state. Today's technology defines a Quantum Robot as a mobile autonomous platform that is equipped with a quantum computer as its processing system (e.g., [3]). Very few references can be found on the applications of quantum capabilities experimentally applied to actual mechanical systems, and to the authors' knowledge, no reference is found that applies experimental quantum capabilities to multibody dynamic systems such as multi-agent robotic problems (for instance, for control and autonomous applications).

The applications of quantum technologies in advancing the performance of mechanical systems (at macro scale) are found in the literature merely on developing novel sensors and actuators ([5]-[10]). A proposed quantum actuator, for instance, is able to manipulate qubits efficiently with described time optimal control sequences [5]. A quantum entanglement-based metrology method has been developed based on a single spin state, using a magnetic resonance approach inspired by the coherent control over multi-body systems [10]. An individual qubit would be used in a quantum sensor to sense the dynamics of its surrounding [11]. Magnetometry techniques have been developed using solid-state qubits [12]. Quantum control techniques are applied, for example, for the reconstruction of the profiles of time-varying magnetic fields [13]. For example, nanoscale magnetic sensing, using coherent manipulation of an individual electronic spin qubit, has been demonstrated experimentally [14]-[15].

The physics community has been investigating quantum information processing systems, with optics-based distributed networks. However, the transfer of knowledge between the physics and engineering communities has been limited [16]. Ultra-low energy optical switching in a cavity quantum electrodynamic system is used for engineering and classical optical models. A nanophotonic approach in building a self-correcting quantum memory, simply "powered" by Continuous Wave (CW) laser beams, has been proposed for developing quantum devices that are able to control themselves [16].

The applications of quantum optics in association with controls [17]-[20], feedback systems [21]-[22], and programmable logic devices in quantum optics [23] can potentially lead to the foundation of the new quantum engineering field for interdisciplinary research. Although there is rich literature in "quantum engineering" ([16]-[23]), the actual integration of such technologies with macro scale mechanical systems as (multi-body) autonomous dynamic systems (e.g., robots, unmanned systems) has not yet been developed to the authors' knowledge.

An experimental quantum-enhanced stochastic simulation device can execute a simulation using less memory than possible by classical means, which integrates experimental quantum interference with soft computing [24].

Free-Space Optical Communications between unmanned systems has also been tested by researchers [25] in a classical sense but not in a quantum context (although quantum mechanics also uses optics and photonics technologies). The main difference between optical control of a multi-agent robotic system and the quantum-enhanced approaches is that the quantum approaches deal with single photon manipulation (alternatively could be electron-based, or even by sound energy levels [26]), which gives advantageous and unmatched capabilities such as the possibility of entangling the robotic agents in a distributed robotic system, quantum superposition, and guaranteed security.

In classical mechanics, the motion of a body is modeled as a particle (point), or as a rigid body. Quantum mechanics is primarily a description of the behavior of elementary particles (e.g., photons, and electrons) at a very small scale. The dynamics of elementary particles can be analyzed by means of the laws of analytical mechanics, applicable to both classical and quantum systems (for example, in [27]). The principles of quantum mechanics applied to rigid body problems can be found in literature. However, no approach is yet available in treating classical multiple-body dynamical systems with quantum mechanics, where the quantum control of the system is of interest. The research by the authors in this paper is the first effort towards the theoretical and experimental research and establishing the interdisciplinary field of Quantum Multibody Dynamics [28] (as an ongoing research activity).

Examples of recent advances towards quantum networks include: Secure quantum communication code, which requires no classical communication [29]; quantum correlations over more than 10 km [30]; Sending entangled particles through noisy quantum channels [31]; Entanglement-based quantum communication over 144 km

[32]; Distributing entanglement single photons through a free-space quantum channel in between cities [33]; and Security of quantum key distribution with entangled photons [34]. A quantum-inspired approach has been proposed to solve problems of two robotic agents finding each other, or pushing an object ([35], [36]), without any knowledge of each other. Furthermore, the research on Psi Intelligent Control inspired by precognition [37] has led to initiating a quantum entanglement-based approach applicable to autonomous vehicle [38].

Recent technological advances make various experimental quantum mechanics accessible with considerable cost and size reductions [39]-[44]. Moreover, reconfigurable Quantum Key Distribution (QKD) networking [45] techniques allow free-space quantum communication over significant distances, and overcomes signal degradation issues (due to weather events, etc.). Perfect security of bit commitment between two mistrustful parties is impossible. However, unconditionally secure bit commitment by transmitting measurement outcomes is possible to attain perfect security when, two agents in a network, Alice and Bob, split into several agents exchanging classical and quantum information at times and locations suitably chosen to satisfy specific relativistic constraints ([46], [47]). Collectively, such technology resources and potential capacities allow us to apply quantum capabilities in engineering applications effectively, particularly in the domains of robotics and autonomy.

The authors introduced briefly the concept of experimental quantum cryptography and entanglement for robotics and autonomy applications in references [48] and [49]. This concept is now discussed and explained in the present paper for the first time as an initiative in an Experimental Quantum Robotics and Autonomy framework. The present paper provides a brief introduction to photons and quantum mechanics in Section 2 (Section 2 intends to allow the readers to use the paper as a self-contained article, particularly for the readers with only a basic background in quantum mechanics). Sections 3 and 4 are dedicated to introduce the interdisciplinary interface of the quantum technology with robotics autonomy, by applying experimental quantum cryptography and entanglement practices to multi-agent robotics and autonomy applications.

# 2 Photons and Quantum Mechanics

This section provides a brief introduction to photons and quantum mechanics. The experimental procedure of a Mach-Zehnder interferometer is presented. The procedure of the experiment is used as the basis of the concepts of experimental quantum photonics. Thehe basis for better understanding of the applied quantum cryptography and entanglement practices associated with the robotics and autonomy applications is presented in Section 3.

## 2.1 Electromagnetic waves

Light, electromagnetic waves, and radiation all contain electromagnetic energy. Light is made of discrete packets of energy called photons. Photons carry momentum, have no mass, and travel at the speed of light. The particle-like nature of light is observed by detectors. Light can be polarized. An electromagnetic signal is a time varying wave that consists of electric and magnetic field components perpendicular to each other and to the direction of the wave propagation (e.g., [50], [51]). In Fig. 1, the electric field, $E_x$, is vertically polarized. There is a corresponding frequency and wavelength associated with electromagnetic energy. Light behavior is realized as wave-like or particle-like. The smallest intensity of light corresponds to a photon. If the intensity is reduced further, it only reduces the frequency of the photon.

A fundamental principle of quantum mechanics is photn superposition (if quantum photonics is under investigation). It can be realized by the spin of electrons [52], or as more recently proposed, based on energy levels of sound-based nanomechanical oscillator [26]). Moreover, if an event can occur in several alternate ways that are indistinguishable, the probability amplitude for such an event is given by the superposition of the probability amplitudes for each event. The quantum mechanics of alternate paths can be summarized in three rules [50]:

1. The probability of an event is given by the square of the absolute value of a complex number $\phi$:
$$P = |\phi|^2$$
where $\phi$ is the probability amplitude of that event.
2. The probability for an event is the sum of the probability amplitudes, if an event can occur in several alternative ways:
$$P = |\phi_1 + \phi_2|^2$$
3. The probability of an event is the sum of the

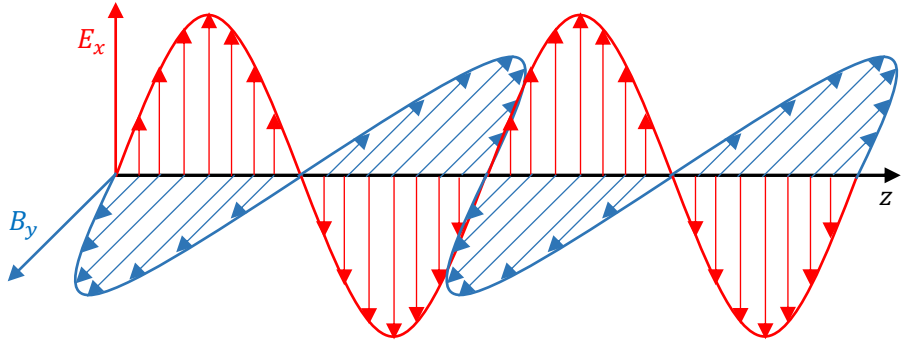

Fig. 1 An electromagnetic wave.

probabilities for each possible alternative, and if the experiment is capable of determining which path is taken:

$$P = |\phi_1|^2 + |\phi_2|^2$$

In a classical sense, a physical system can have multiple possible states, where the states may denote physical situations of the system with different measurable properties. Measurements of a physical property of the system in each of the states are independent of the others. In a quantum sense, a quantum of light (i.e., a photon) is considered as the system. Considering the Mach-Zender interferometer example, if an incident photon enters from A in Fig. 2, there are two possibilities to go from A to B (or C). Taking path $l_1$ can be associated with the state of the system being in State 1, and taking path $l_2$ can be associated with the state of the system being in State 2. Therefore, the paths can represent the state of the system. In quantum mechanics, unlike in classical systems, the system can be in superposition of the two states simultaneously, which is non-intuitive in a classical sense. There is a 50% chance that a photon incident entering the beam-splitter at A reflects, and 50% chance for it to be transmitted. The paths in this figure can represent the state of the system (photon). If we cannot distinguish the paths, or which path the photon takes, then the paths are called indistinguishable. If we block one path, for example by a photon trap, then we know that the photon can only go through the other path in going from A to B (or C). In this case, the paths are distinguishable, as we know that there is only one possible path that the photon can take. When the paths are indistinguishable, the state of the photon will be in superposition of both paths. Quantum mechanics predicts that a system can be in two distinct states at the same time. By measurement, one can find the state of the system. However, a peculiarity of quantum mechanics is that the state of the system cannot be predicted before the measurement is made. Quantum entanglement predicts non-local behavior where two particles can be entangled. Pairs of photons can be created simultaneously by the spontaneous parametric down-conversion (SPDC) procedure, experimentally. SPDC produces photon pairs that are entangled in polarization. This process is discussed in the entanglement section of the present paper.

In the Mach-Zehnder interferometer experiment in Fig. 2, if the paths are indistinguishable, the probability amplitude of the photon in going from A to B is the superposition of the probability amplitudes in going through each arm, $\phi_1$ and $\phi_2$.

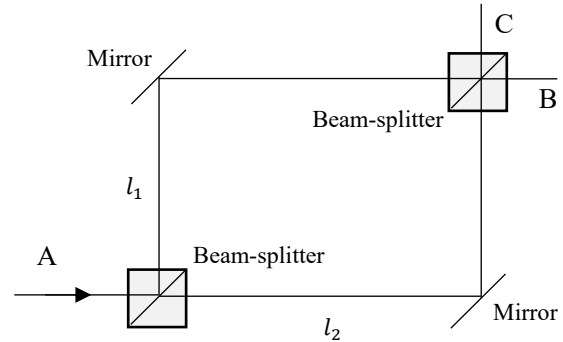

Fig. 2 A Mach-Zehnder interferometer.

The beam-splitters in Fig. 2 reflects half of the intensity of the incident light (entered form A) and transmits half of the intensity. Therefore, the probability of a photon being reflected or transmitted is $\phi=1/2$. The probability amplitude of being reflected or transmitted is the square root of the probability, which is $1/\sqrt{2}$. If arm 2 is blocked, the probability amplitude of the photon being reflected at the first beam splitter is $1/\sqrt{2}$, and the probability amplitude of the transmitting at the second beam splitter is also $1/\sqrt{2}$. The probability amplitude for the photon in going from A to B is computed by the product of the two probability amplitudes, or $|\phi_1| = 1/\sqrt{2}$. $1/\sqrt{2}$=1/2. When arm 2 is blocked, the probability of the photon going from A to B is $P_1 = |\phi_1|^2 = 1/4$. If the arms are not blocked, there are two different probability amplitudes. If the paths are indistinguishable, the superposition of the sum of the probability amplitudes gives

$$\phi = \phi_1 + \phi_2$$

Thus, the probability of the event for a photon going from A to B, in case of two paths, is obtained by adding the probability amplitudes $|\phi_1|$, and $|\phi_2|$ as,

$$P = |\phi|^2 = |\phi_1|^2 + |\phi_2|^2 + 2|\phi_1||\phi_2|\cos\delta \quad (1)$$

where $\delta_1 = 2\pi/\lambda$, and $\delta_2 = 2\pi/\lambda$ are the phases corresponding to $|\phi_1|$ and $|\phi_2|$ probability amplitudes, respectively, as shown in Fig. 3, and $\delta = \delta_2 - \delta_1$.

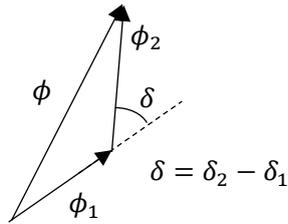

Fig. 3 Adding the probability amplitudes.

In the Mach-Zehnder interferometer when $|\phi_1|=|\phi_2|=1/2$, from Equation (1) we obtain

$$P = 1/2(1 + \cos\delta) \quad (2)$$

For, $\delta = 2n\pi$, where $n = 0,1,2,3,...$, the probability of a photon going from A to B is 1, or $P = 1$, which means that every photon will reach B. For, $\delta = n\pi$, where $n = 1,3,5,...$, the probability of a photon going from A to B is 0, or $P = 0$, which means no photon reaches B.

Single photon detection may be achieved by sending a very weak beam of light to a detector (e.g., a photon counter). By sending $N$ photons in one second, the detector will record $NP$ photons, where $P$ is the probability in Equation (2).

In the interferometer in Fig. 2, the photon is subject to superposition of taking the two paths, and it interferes with itself. Increased intensity of light causes many photons (of the order of many billions), to pass through the interferometer, where each photon interferes with itself.

The SPDC process can generate pairs of photons simultaneously. In this process one photon is split into two when passing through a nonlinear crystal (Fig. 4).

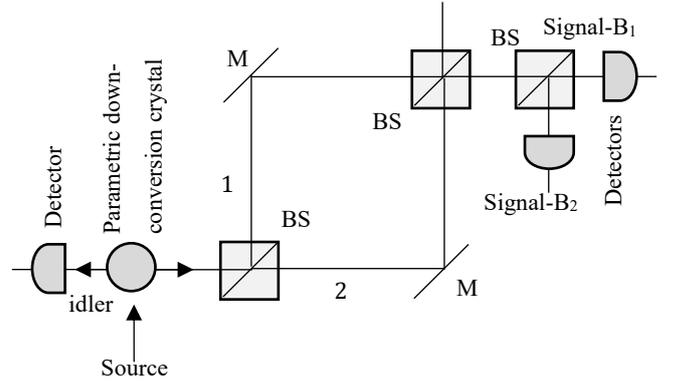

Fig. 4 Parametric down-conversion setup, and Mach-Zehnder interferometer.

In a parametric down-conversion process, detection of one incident photon is required. A very weak laser light source (about $10^6$ photons/s) can be used for the single photon generation purpose. The weak source can give a close enough correlation with the single photon generation in an experiment.

In Fig. 4, a photon is split into two, by a nonlinear crystal, in the SPDC process. The spliting of photons in SPDC in fact is the process that takes place on the nonlinear crystal, where one photon incident is converted into two photons with lower energies (that add up to the energy of the parent photon). One photon, called the idler, is sent to a detector, and the other, called the signal, is sent to the interferometer, and finally to a detector. If one photon is sent to the interferometer, we only know the probability of which arm (1, or 2) it will go through. This is a characteristic of quantum mechanics; specifically, quantum mechanics is probabilistic and classical mechanics is deterministic.

In Fig. 5, a Beam Block or Trap is placed along Arm 2. This makes the paths of the interferometer distinguishable (whereas, before placing the Block, the paths were indistinguishable).

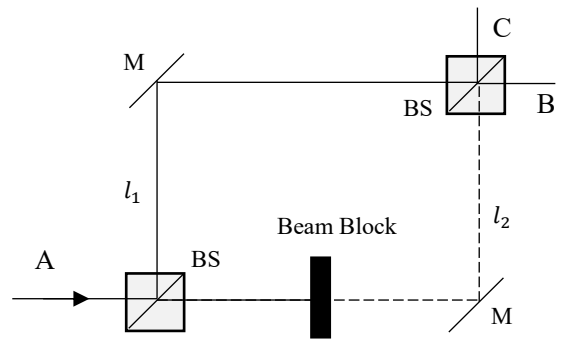

Fig. 5 A Mach-Zehnder interferometer with a distinguishable path.

When one arm is blocked, there is only 25% probability that the photon will reach detector B. A photon that reaches detector B may be interpreted as "the photon knows that there is block in arm 2" and it detects the block without going through it. This quantum prediction in such experimental scenario is impossible in a classical mechanics sense [50].

## 2.2 Plane waves and polarization

A plane wave is a constant-frequency wave with its wave-fronts, or surfaces of constant phase, as parallel planes of constant amplitude waves traveling normal to the phase velocity vector. The equation of the displacement $y$ of a plane wave, travelling along the $x$ axis, can be given by

$$y = A \cos(\frac{2\pi x}{\lambda} - \frac{2\pi t}{T}) \quad (3)$$

where $A$ denotes the amplitude of the wave, $\lambda$ is the wavelength, and $T$ is the period of oscillation. In this equation, for any $x$, and $t$, the waves on the $yz$ planes, (perpendicular to $x$ axis), have the same phase.

Electromagnetic waves are transverse waves where the direction of the oscillation is perpendicular to the direction of travel. If a light wave travels along the $z$ axis, with the electric field vectors in the $y$ direction, the wave equation can be expressed by

$$E(z,t) = (0, E_0, 0) \sin\left(\frac{2\pi z}{\lambda} - \frac{2\pi t}{T}\right) \quad (4)$$

where $E_0$ denotes the amplitude, and $E$ is the electric field at position $z$, and time $t$. When the electric field vectors are in the $y$ direction, the waves have the same phase along $z$ in each $xy$ plane. Transverse waves can be polarized. Polarized waves in an electric field always point parallel to the same direction. A wave polarized along the $x$ axis is called a linearly polarized wave in the $x$ direction.

The electric field vectors can be resolved into orthogonal components. For waves traveling in one dimension along an axis, the electric field can be considered in a vertical plane (Fig. 1). If the wave is propagating along an axis perpendicular to the plane of the paper, the horizontal ($H$) and vertical ($V$) orthogonal axes can be used to describe the electric field (Fig. 6).

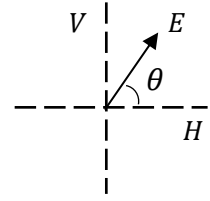

Fig. 6 Electric field of light.

If the horizontal-vertical ($HV$) axes are rotated by angle $\theta$, we have a new set of axes, $H'V'$. An electric field of light with an amplitude of $E_0$ can be represented in the rotated axis as in Fig. 7, where the amplitude, $E_0$, resolved along $H'V'$ axes gives $E_0 \sin\theta$, and $E_0 \cos\theta$.

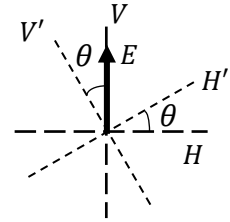

Fig. 7 Electric field of light.

## 2.3 Polarized photons

Light is unpolarized, or in another words it includes polarizations with various orientations. Light is linearly polarized when passing through polarizing filters. Polarizers can align the oscillating field in one direction. A polarizer absorbs light parallel to an internal axis, known as the extinction axis, and transmits incident light parallel to the orthogonal axis, known as the transmission axis. Fig. 8 illustrates a polarizer with its transmission axis oriented vertically, which polarizes the light vertically (shown by arrows) when passing through it, for a travelling wave perpendicular to the plane of the paper.

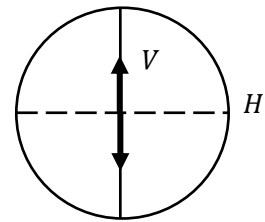

Fig. 8 Vertically polarized light passing through a vertical polarizer.

If the transmission axis of the polarizer is perpendicular to the oscillation of the incident electric field, as shown by the solid line in Fig. 9, then the light is absorbed and no light passes through the polarizer.

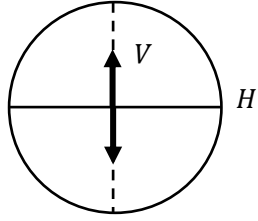

Fig. 9 Vertically polarized light is absorbed in a horizontal polarizer.

If vertically polarized light enters a polarizer with its transmission axis oriented at an angle, $\theta$, as shown in Fig. 10, a component of the incident light passes through the polarizer, and the rest is absorbed.

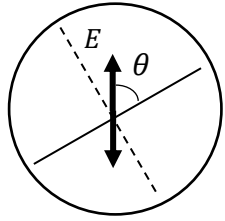

Fig. 10 Vertically polarized light through a polarizer with its transmission axis aligned with angle $\theta$ relative to the vertical axis.

The component of the incident light that passes through the polarizer is $E_0 \cos\theta$. The component $E_0 \sin\theta$ is absorbed by the polarizer. Therefore, the transmitted electric field is $E_T = E_0 \cos\theta$. Based on Malus' law, the transmitted intensity is $I_T = I_0 \cos^2\theta$. In a quantum sense, the above discussion on polarization is valid; however, at the photon level.

As discussed before, the probability of the result of an experiment or analysis in quantum mechanics is the square of the probability of the corresponding amplitude. For photons initially polarized with angle $\theta$ relative to the horizontal direction, the probability of detecting the photons after passing through a horizontal polarizer is $\cos^2\theta$ for the incident photons, and this probability is $\sin^2\theta$ when the polarizer is vertical. In this scenario, the photon is said to be in a superposition of horizontal and vertical polarization, with amplitudes $\cos\theta$, and $\sin\theta$, respectively. In summary, if a photon is sent to a polarizer, with polarizing transmission axes of angle $\theta$ relative to the orientation of the photon, the amplitude of the transmission of the photon is $\cos\theta$, with probability of transmission of the photon as $\cos^2\theta$. Vertically polarized photons are not transmitted through a horizontal polarizer. However, if a polarizer with orientation angle $\theta$ relative to the horizontal direction is placed before the horizontal polarizer, some photons are transmitted.

This brief background to quantum mechanics is given as an introduction for experimental quantum cryptography and entanglement, presented in Section 3, in conjunction with the Quantum Robotics and Autonomy problems.

# 3 Quantum Cryptography for Robotics and Autonomy

Experimental quantum cryptography is carried out by polarizing photons, passing them through a beam splitter cube, and detecting the photons' polarizations. The main operations in a quantum cryptography experiment are as follows:

- Sending single photons by a laser diode, and using the Spontaneous Parametric Down-Conversion (SPDC) technique.
- Polarizing photons: performed by sending photons through a $\frac{\lambda}{2}$ (half-wave) plate, where the plate rotates the polarization of the incident light by twice the rotation of the plate (in a physical sense, but here we only refer to the result of the final polarization rotation, and not the plate rotation).
- The process of detecting the polarizations: the polarized photons are sent to a beam splitter (BS) cube, where the BS passes through the horizontal polarizations, and reflects by vertical polarizations. Once the horizontally or vertically polarized photons are passed or reflected by the BS, two sensors, each dedicated to vertical or horizontal polarization sensing, respond to the receiving of the photons.

The experimental setup is illustrated in Fig. 11. The corresponding experimental quantum cryptography components assembled on a mobile robotic platforms are shown in Fig. 12.

For the laboratory demonstration of experimental quantum cryptography, we assume that the laser diode produces single photons, by applying the SPDC process, and single photons are detected by the sensors. However, this is only an assumption, and SPDC is not performed, and Single Photon Counters (SPC) are not used in Fig. 12 (these processes however are presented in Section 4). For demonstration purposes only, instead of single photons, laser pulses are used, which is sufficient for the basic demonstration of the concept of experimental quantum cryptography in the present paper. SPDC and SPC are discussed in the next section, in the experimental quantum entanglement.

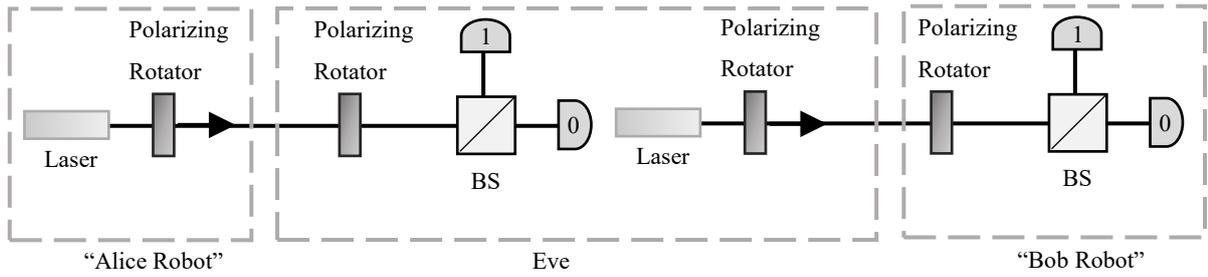

Fig. 11 A diagram for Experimental Quantum Cryptography.

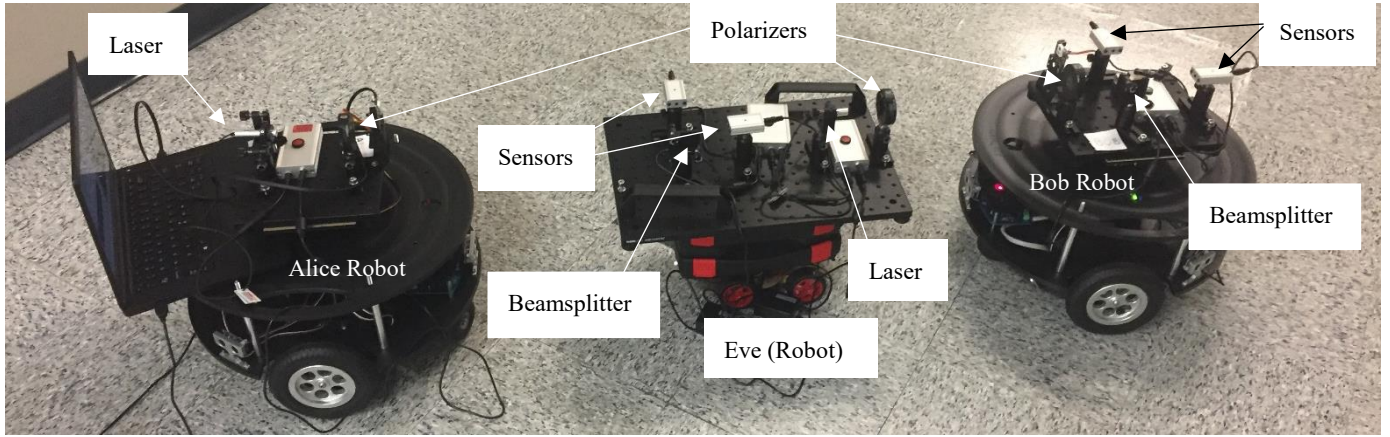

Fig. 12 Quantum cooperative ground robots.

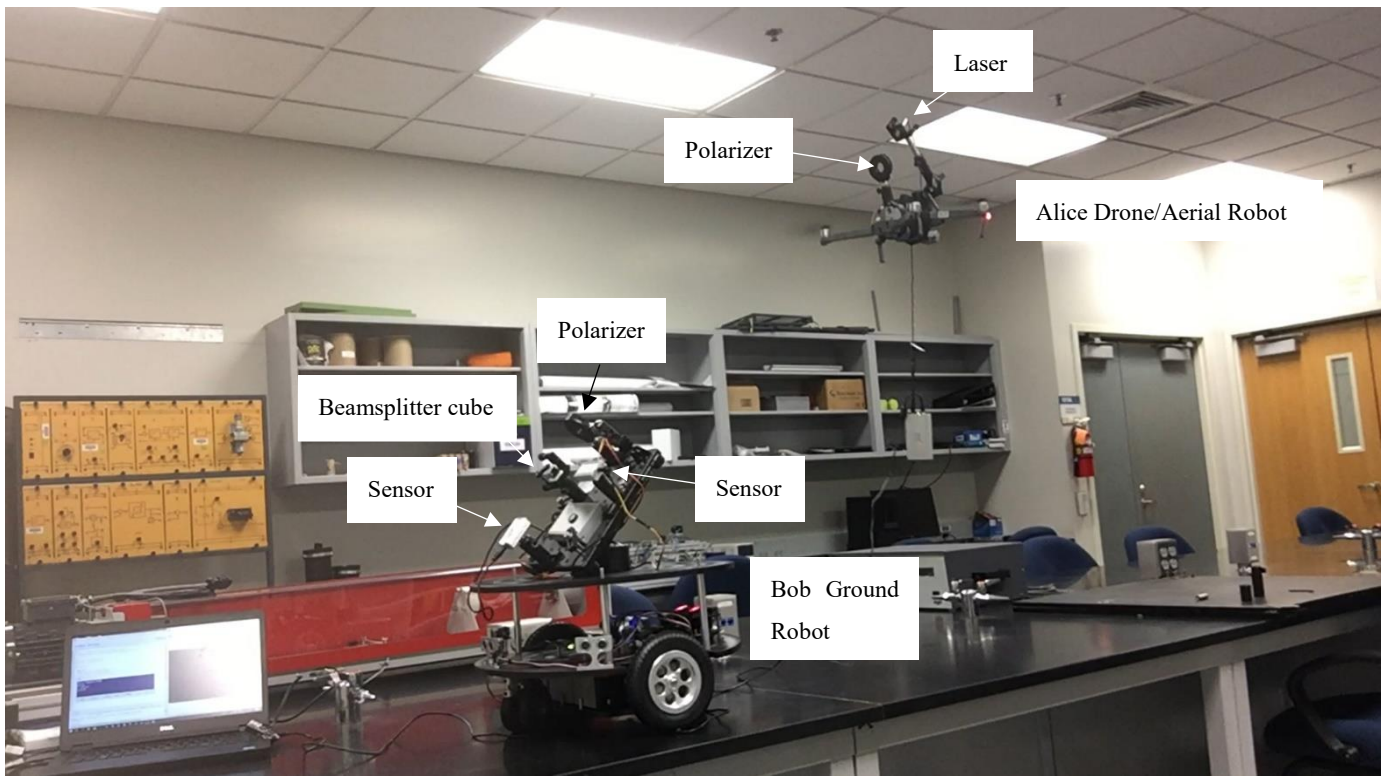

Fig. 13 Quantum cooperative aerial and ground robots.

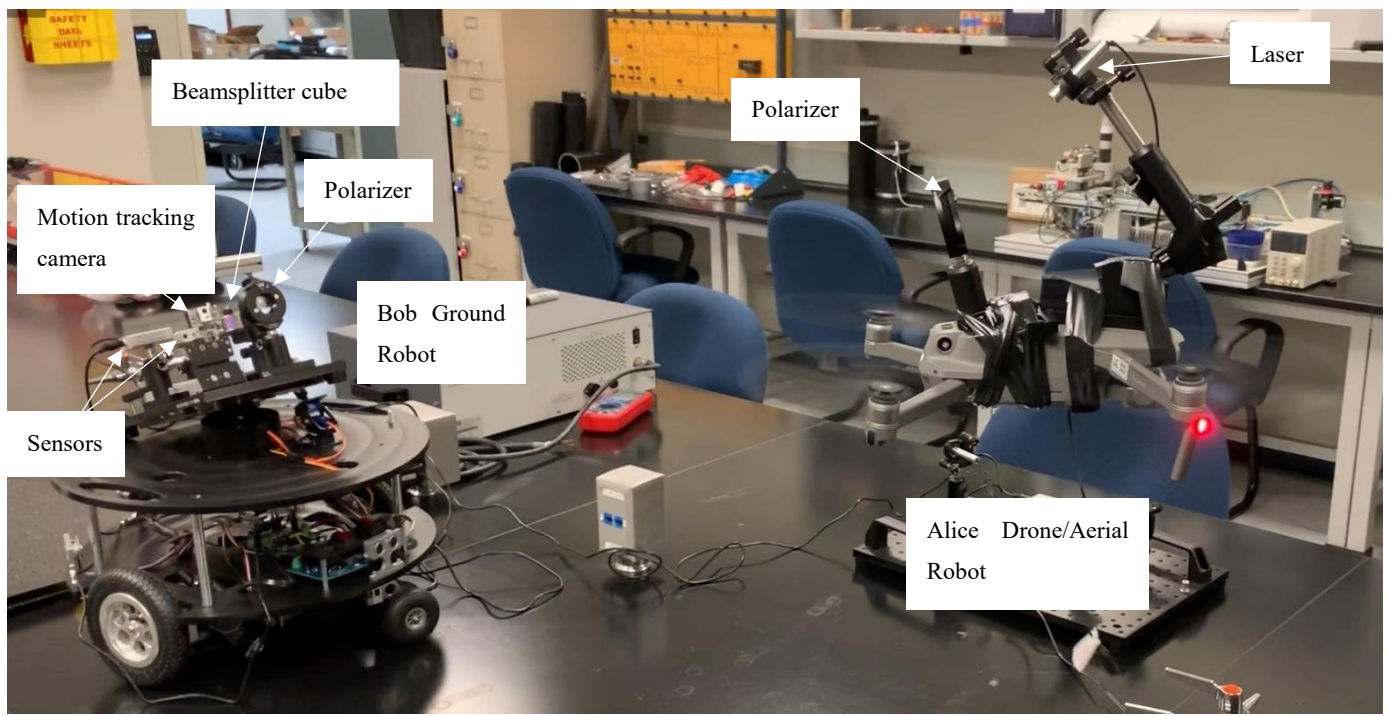

Fig. 14 Quantum cooperative aerial and ground robots.

Producing and detecting/counting single photons is presented in the entanglement experiment in Section 4, and is used in any quantum experiment (rather than laser light pulses, which are only used for demonstration purposes in this section). It should be noted that both generating single photons, and the SPC, are used for the quantum cryptography experiment. The mobile robots in conjunction with quantum cryptography tools are used for collaborative/cooperative tasks and control of the mobile robot applications. The experimental setup is illustrated in Fig. 11. "Alice robot" contains a laser diode, and a $\frac{\lambda}{2}$ polarizing rotator plate. "Bob robot" contains a $\frac{\lambda}{2}$ polarizing rotator plate, a beam splitter cube, and two sensors. Eve, eavesdrop attacker, tries to a) intercept and detect the information sent from Alice, and b) duplicate the information and transfer it to Bob, so that Alice and Bob cannot notice the presence of the eavesdrop. However, Eve is exposed to Alice and Bob after few exchanges of photons, for two fundamental reasons, which make it impossible for Eve to intercept. The first reason is that, a single photon cannot be partially detected (as a part of a laser pulse can, for example), and once a single photon is detected by a sensor, it is actually trapped. The second reason is that polarization of a single photon (e.g., vertical and horizontal) can be produced by a variety of combined polarizations. When a photon passes through two separate polarizers, one on Alice, and one on Bob, the result of the two polarizers are combined. So, by having two polarizers (one on Alice, and one on Bob), there are various ways of combining polarizations to achieve one result. This is discussed next.

The "ket" symbol $|\ \rangle$ is used to denote a quantum state. The combination of polarizations associated with the half-wave plates at Alice and Bob are given in Table 1 ([53]), where "+" basis corresponds to $|0º\rangle$, and $|90º\rangle$, polarizations (or horizontal and vertical polarizations) and "×" basis corresponds to $|-45º\rangle$ and $|45º\rangle$ polarizations produced by the half-wave plate polarizers. Although Alice and Bob do not exchange any information publicly about the polarization states ($|-45º\rangle$, $|0º\rangle$, $|45º\rangle$, and $|90º\rangle$), both Alice and Bob publicly share the information about which bases ("+" or "×") they use. As shown in Table 1, combination of the + bases of Bob and Alice half-wave plates gives binary 0 and 1 results. Similarly, combination of the × basis of Bob and Alice half-wave plates gives binary 0 and 1 results. These 0 and 1 binary results, which are produced by the combined polarization of the photons going through both Alice and Bob polarizers, will be utilized as control commands in the autonomous operation of the robotic platforms. Fig. 12 shows Alice, Eve, and Bob on mobile ground robotic platforms. Motion tracking cameras on mobile robots are used to keep the laser incident from Alice aligned with Bob's polarizer, beam splitter, and sensors at all times. The motion tracking cameras activate servo motors, which are placed in between the optics equipment and the mobile platforms for the alignment purpose. In Fig. 13, Alice is an aerial robot and Bob is a ground mobile robot. The same quantum cryptography process as in Fig. 12 is used in the experiment in Fig. 13. For aligning the laser incident in a three dimensional problem, in the cases for multi-agent aerial/ground robots or aerial/aerial robots, and the cooperation of ground/ground robots (when there is any relative elevation change between robots), each robot includes two servo motors in order to provide yaw and tilt tracking degrees of freedom (Fig. 13 and Fig. 14). Fig. 14 provides a different view of the same aerial/ground robot collaborative task demonstration of Fig. 13, for better visualization of the experiment. Thorlabs equipment [53] was used for all the quantum related experiments in the present paper. Arlo robots [54] are used for mobile ground platforms. DJI Mavic 2 [55] is used for the drone experiment [55]. Pixy2 cameras are used for motion tracking (while finer alignment is required for better results, we will improve the system in a future work). The description of the fine tuning of the alignment of the optics equipment, with the laser light incidents on one mobile robot aligned with the polarizer on another robot, can be found in other references such as [24].

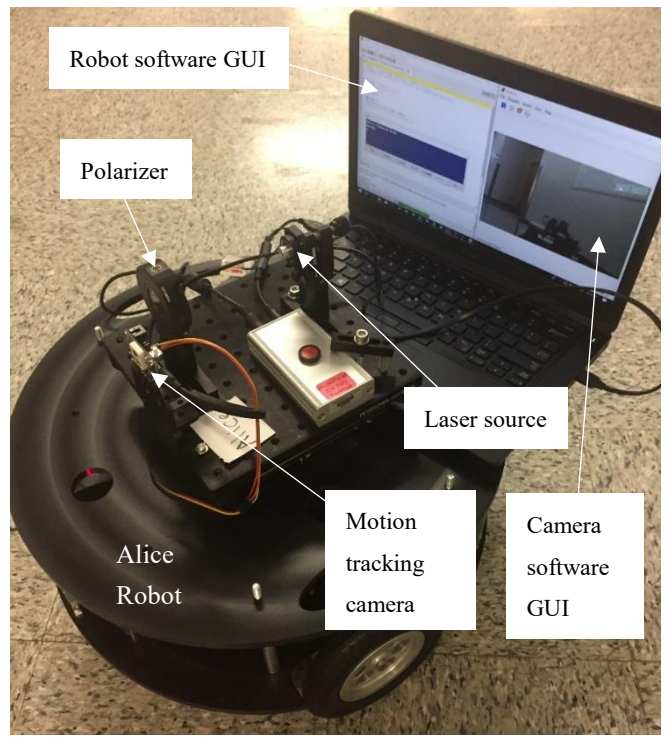

Fig. 16 Alice ground robot (with the mobile robot platform software, and motion tracking camera software running on a laptop).

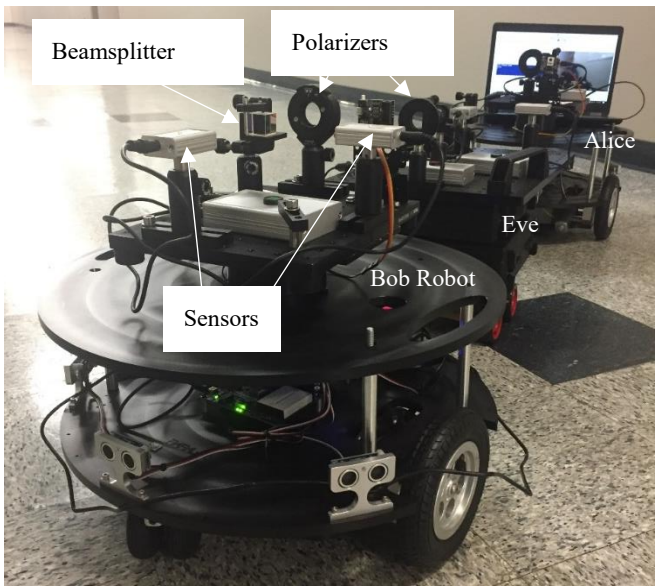

Fig. 15 Bob ground robot.

Fig. 15 and Fig. 16 show Bob and Alice, respectively, and Fig. 17 and Fig. 18 show Alice drone in different views (for providing better visual presentations of the experiments).

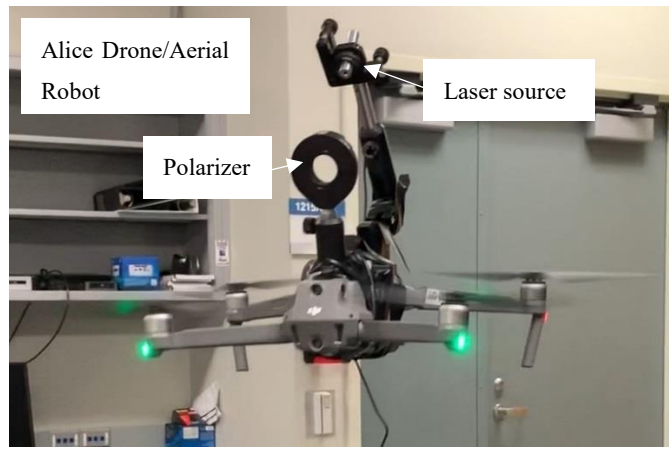

Fig. 17 Alice drone/aerial robot.

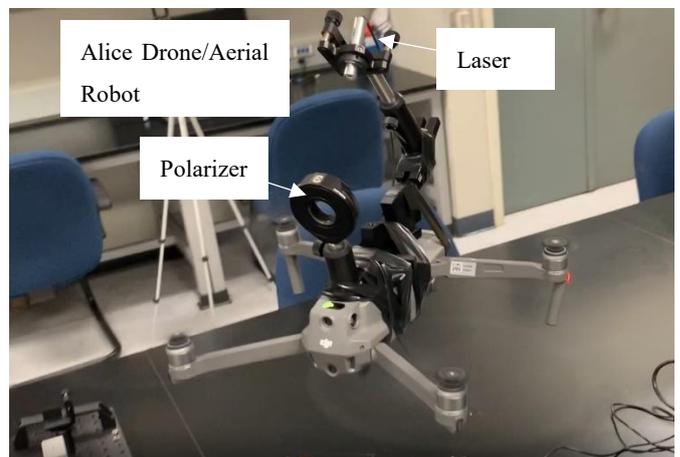

Fig. 18 Alice drone/aerial robot.

Table 1. Polarizing combinations associated with the half-wave plates at Alice and Bob.

| Alice (Half-wave plate polarizer in degrees) (BASIS: + or ×) | Bob (Half-wave plate polarizer in degrees) (BASIS: + or ×) | Result | Beam Splitter cube | Binary result (Sensor 0 or sensor 1) |
|---|---|---|---|---|
| $\|0°\rangle$ (+) | $\|0°\rangle$ (+) | $\|0°\rangle$ | Passes the light | 0 |
| $\|90°\rangle$ (+) | $\|0°\rangle$ (+) | $\|90°\rangle$ | Reflects the light | 1 |
| $\|45°\rangle$ (×) | $\|0°\rangle$ (+) | Random $\|0°\rangle$ and $\|90°\rangle$ | 50% reflects 50% passes light | No result (Discard) |
| $\|-45°\rangle$ (×) | $\|0°\rangle$ (+) | Random $\|0°\rangle$ and $\|90°\rangle$ | 50% reflects 50% passes light | No result (Discard) |
| $\|0°\rangle$ (+) | $\|45°\rangle$ (×) | Random $\|0°\rangle$ and $\|90°\rangle$ | 50% reflects 50% passes light | No result (Discard) |
| $\|90°\rangle$ (+) | $\|45°\rangle$ (×) | Random $\|0°\rangle$ and $\|90°\rangle$ | 50% reflects 50% passes light | No result (Discard) |
| $\|45°\rangle$ (×) | $\|45°\rangle$ (×) | $\|90°\rangle$ | Reflects the light | 1 |
| $\|-45°\rangle$ (×) | $\|45°\rangle$ (×) | $\|0°\rangle$ | Passes the light | 0 |

When an eavesdropper "Eve" intercepts the signal between Alice and Bob, it detects Alice's polarization as 0 or 1, but cannot distinguish if it has been produced by polarization combinations of $|-45°\rangle$ and $|45°\rangle$, or $|0°\rangle$ and $|0°\rangle$. Therefore, when Eve tries to duplicate the message (polarization) that is produced by Alice, there is only a 50% chance that Eve can send to the correct polarization. Alice and Bob compare their bases, for a number of bits that has been already exchanged between them. This exchange of information is public, and does not need to be encrypted. If after comparison, the bases do not match, then this error translates into the presence of Eve intercepting. Because there is a chance that Eve can predict the correct polarizations, by comparing more transferred bits by Alice and Bob, there is a larger chance of detecting the eavesdropper's interception.

We can now translate the binary results in Table 1 simply to local digital commands for the robots. The autonomous mechanism of the motion for the mobile robotic platforms can now be designed by using the binary results of the quantum cryptography as inputs to the onboard digital microcontrollers. The controller is programmed to output motion commands to the robot actuators (e.g., electric motors) on the mobile platforms. For example, the simplest form of command can be defined by:

- *If the binary result is 1, then the command is: rectilinear translation of the mobile platform with a predefined constant velocity.*
- *If the binary result is 0, then the command is: velocity of the mobile platform is equal to zero.*

The binary results of the experimental quantum cryptography process can define any desired digital protocol. In this way, specific robotic tasks can then be identified, and designed for the corresponding desired case studies.

# 4 Quantum Entanglement for Robotics and Autonomy

Quantum entanglement occurs when groups of particles/agents interact in ways such that the quantum state of each particle cannot be described independently of the state of the others. Pairs of photons that are entangled in polarization can be generated simultaneously by the Spontaneous Parametric Down-Conversion (SPDC) process. Quantum entanglement predicts non-local behavior, where two particles are entangled. Photons can be in two states of vertical or horizontal linear polarization. Entanglement can be specified by two photons being in orthogonal linear polarizations. In such polarization entanglement, one photon can exhibit vertical polarization, and the other photon horizontal polarization, where each photon can be in superposition of being vertically or horizontally polarized. The polarization of each photon being vertical or horizontal remains unknown until a measurement is made. Quantum mechanics only predicts that the photons are in vertical and horizontal polarization states, simultaneously, but the state of the polarization of the photons cannot be individually labeled for each photon. When a measurement is made, we can find each of the two photons in one of two states with a corresponding probability. Measurement of the polarization reveals the polarization of a photon being vertical or horizontal. By knowing the polarization of one photon, the polarization of the other photon can be predicted as the polarizations of the photons are orthogonal in the SPDC process. However, the two photons are only in entangled states until the moment the measurement is made. Once the measurement is made, the two photons will no longer be entangled. Therefore, the entangled photon pairs only remain correlated before a measurement is made. In summary, a pair of photons can be entangled in polarization, which is a non-local property of quantum mechanics (Violation of Bell's inequalities proves non-locality [57]), but we do not know the polarization of each photon, and when we measure the polarization of one, they are no longer entangled. Non-local properties have led to many remarkable results and applications such as quantum teleportation [58], and the rise of the new field of quantum information (e.g., [59]-[61]).

A quantum entanglement experiment is shown in Fig. 19 (Notations in the figure are: (M: Flipper mirror, M: Mirror; MA: Mirror A; MB: Mirror B; HWP: Half-wave plate; AP: Autonomous (Mobile) Platform). The physical system is presented in Fig. 20. SPDC process is used to convert one photon of higher energy into a correlated pair of photons with lower energies (where the energies of the correlated pair add up to the energy of the parent photon). By sending a violet pump laser beam (e.g., a 100 mW laser, with 405 nm wavelength) through a nonlinear crystal (e.g., BBO: beta Barium Borate), one photon of higher energy is converted into a correlated pair of photons (with 810nm wavelength), producing entangled states, each having correlated photon with the same energy, with orthogonal horizontal and vertical polarizations. The two single photon counter (SPC) detectors, in Fig. 21 (SPCM50A/M [53]), installed on two robots detect the probability measure of the photon incidents. They arecounted by the photon detectors and used for evaluating entanglement. The SPCM50A/M module specifications include: Wavelength Range of 350 - 900 nm, Typical Max Responsivity of 35% at 500 nm, and Active Detector Size 50 μm ([53]). The 810 nm narrow bandpass filters, with a Bandwidth of 30 nm, shown in Fig. 21, will allow only 810 nm photon pairs produced by the SPDC process to reach the SPCs.

The process of identifying the entangled correlated photon pairs entails the detection of the photon pairs that reach the two SPC detectors at the same time [64]. The signals from the two SPC detectors can be sent to an electronic coincidence unit. Each SPC detector assigns a coincidence to any pair of detected pulses that arrive within a specified time (basically an AND gate). Another possibility could be to have a circuit that assigns a time for the arrival of the photons, so that through a classical channel Alice and Bob can compare the arrival times, and those that arrive within a specified time can be considered in coincidence. The SPC detectors on Alice and Bob should synchronize their clocks to within tens of nanoseconds or perhaps pick the time from wireless signal, and then save the time of arrival of the detector photons. Alice and Bob already need to share the basis information through the classical channel, anyway (e.g., BB84 encoding), so they can share the photon arrival times as well [64]. If the HWP and the polarizer in Fig. 19 is used to manipulate the polarization state of one photon, the corresponding photon pair is effected simultaneously as the result of the entanglement phenomenon, which retains the photons in an entangled state (until before measurement of a photon state is made).

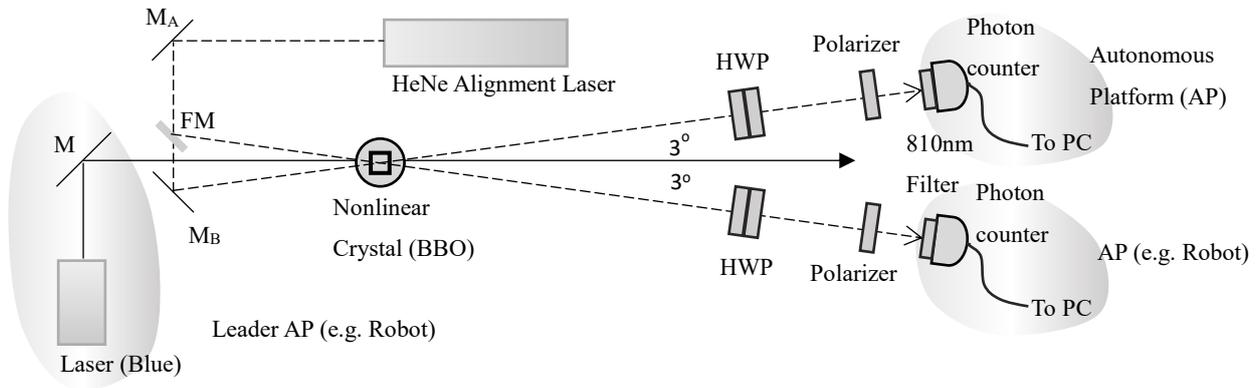

Fig. 19 Entangled robots.

When entangled photon pairs are detected by the single photon counters, a digital signal from the counters can be sent to the onboard microcontroller of a robot, which then can be translated into a digital task for control of the robot actuators. Once Alice and Bob robots are in an entangled state, specific information can be exchanged (e.g., by quantum cryptography process) between a leader robot and the entangled robots, for the two robots (Fig. 19 and Fig. 20) to perform simultaneous robotic/autonomous identical tasks.

The quantum collaborative autonomous platforms presented in the cryptography and entanglement section of the present paper, and also a combination of scenarios of entanglement and cryptography for autonomy, (where entanglement triggers the process of Cryptography for multiple robotic systems) perhaps can be the most sophisticated technique in cooperative robotics and unmanned systems technology. This is due to the ultimate speed of photon propagation for robotic control applications, truely guaranteed security and immunity against cyberattacks, and the possibility of having access to the entanglement capabilities (which does not exist in the classical domain).

As suplimental material, some videos of the quantum robotic experiments, presented in the present paper, are available in References [62] and [63].

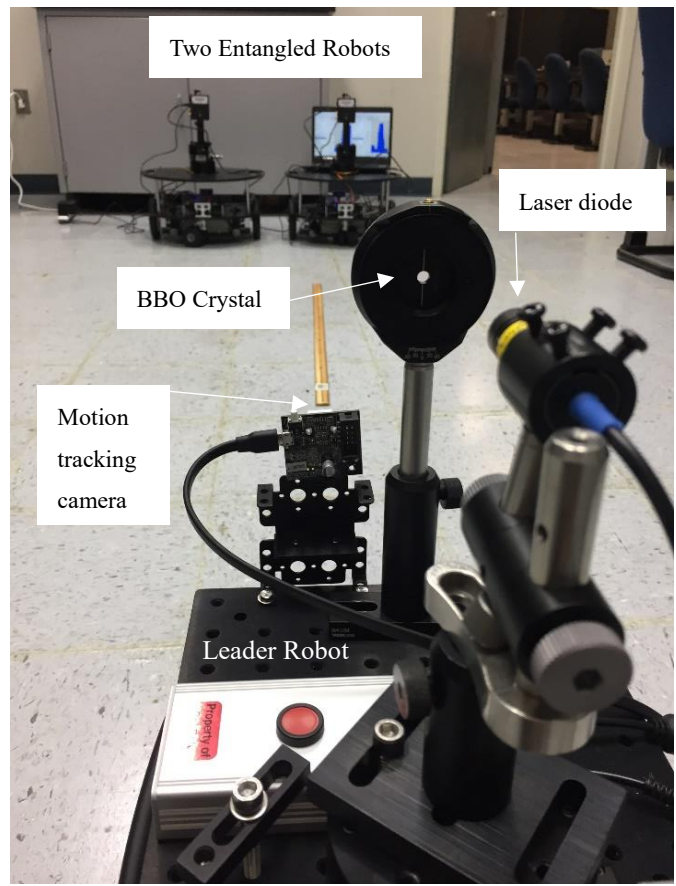

Fig. 20 Entangled robots experiment.

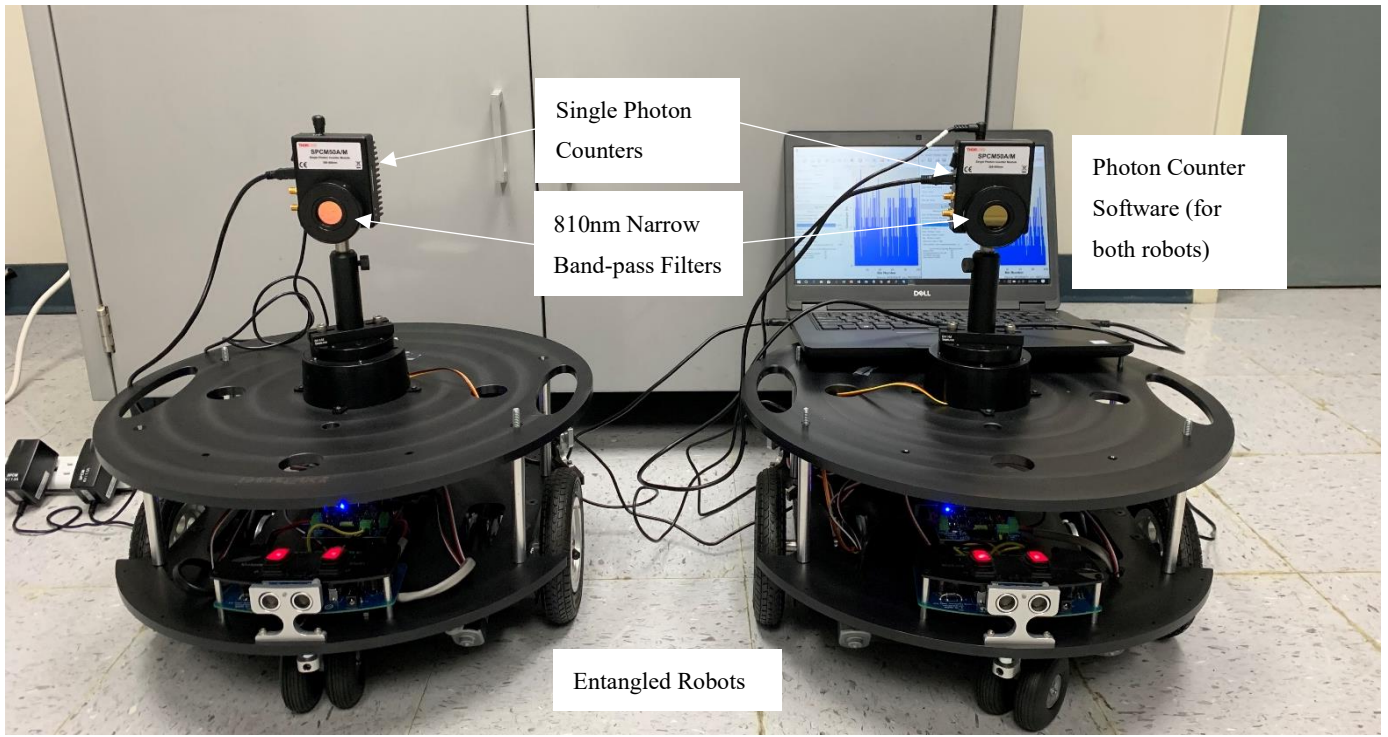

Fig. 21 Entangled robots, single photon counters (with the single photon counting software in the background).

## 5 Conclusions

A quantum network of unmanned autonomous vehicles or robots can potentially offer significantly superior capabilities in comparison to classically networked dynamical system. This is due to accessing the speed of photon propagation or speed of light, which allows achieving the ultimate communication speed for control purposes, guaranteed immunity against cyberattacks, and unmatched quantum entanglement capability (which does not yet exist in the classical domain of robotic autonomy). Experimental quantum cryptography and entanglement for cooperative robotics applications were introduced in the present paper. A scenario where quantum entanglement can be achieved in practice using correlated photon pairs was introduced for placing two robots in an entangled state. Experimental quantum cryptography process in the presence of eavesdrop intercept for robotics applications was introduced. The intent of the present paper was to provide the pioneering idea of Quantum Cooperative Robotics. Therefore it attempted to provide an introduction for further investigations and a catalyst for open discussions. The authors are now developing a generalized framework based on this article on the topic of Quantum Multibody Dynamics [28] where the classical kinematics and kinetics of multiple rigid bodies and particles are leveraged in a quantum physical context.

**Acknowledgement**

Government sponsorship is acknowledged. Dr. Quadrelli's contribution was carried out at the Jet Propulsion Laboratory, California Institute of Technology, under a contract with the US National Aeronautics and Space Administration (NASA).

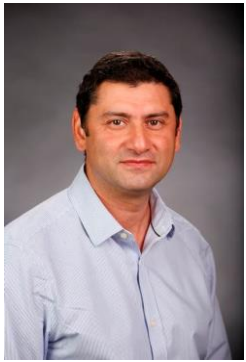

**Farbod Khoshnoud**, PhD, CEng, PGCE, HEA Fellow, is a faculty member in the college of engineering at California State Polytechnic University, Pomona, and a visiting associate in the Center for Autonomous Systems and Technologies, and Aerospace Engineering at California Institute of Technology. His current research areas include Self-powered and Bio-inspired Dynamic Systems; Quantum Multibody Dynamics, Robotics, Controls and Autonomy, by experimental Quantum Entanglement, and Quantum Cryptography; and theoretical Quantum Control techniques. He was a research affiliate at NASA's Jet Propulsion Laboratory, Caltech in 2019; an Associate Professor of Mechanical Engineering at California State University; a visiting Associate Professor in the Department of Mechanical Engineering at the University of British Columbia (UBC); a Lecturer in the Department of Mechanical Engineering at Brunel University London; a senior lecturer at the University of Hertfordshire; a visiting scientist and postdoctoral researcher in the Department of Mechanical Engineering at UBC; a visiting researcher at California Institute of Technology; a Postdoctoral Research Fellow in the Department of Civil Engineering at UBC. He received his Ph.D. from Brunel University in 2005. He is an associate editor of the Journal of Mechatronic Systems and Control.

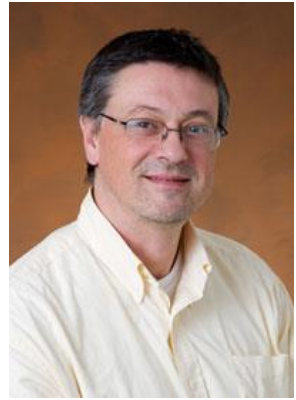

**Dr. Marco B. Quadrelli** is a Principal and Group Supervisor of the Robotics Modeling and Simulation Group at the Jet Propulsion Laboratory (JPL), California Institute of Technology. He received the Laurea in mechanical engineering from the University of Padova, a M.S. in aeronautics and astronautics from MIT, and a Ph.D. in aerospace engineering from Georgia Institute of Technology in 1996. He has been a Visiting Scientist at the Harvard-Smithsonian Center for Astrophysics, did postdoctoral work in computational micromechanics at the Institute of Paper Science and Technology, and has been a Lecturer in aerospace engineering at both the JPL and at the California Institute of Technology Graduate Aeronautical Laboratories. He is a NASA NIAC Fellow, a Keck Institute for Space Studies Fellow, and an AIAA Associate Fellow. His flight project experience includes the Cassini–Huygens Probe Decelerator; Deep Space One; the Mars Aerobot Program; the Mars Exploration Rover and Mars Science Laboratory Entry, Descent, and Landing; the Space Interferometry Mission; and the Laser Interferometry Space Antenna. He has been involved in several research projects in the areas of flexible multibody dynamics; tethered space systems; distributed spacecraft and robots; active granular media; hypersonic entry and aeromaneuvering; planetary sampling; integrated modeling of space telescopes and inflatable spacecraft; and surface vessel dynamics and state estimation.

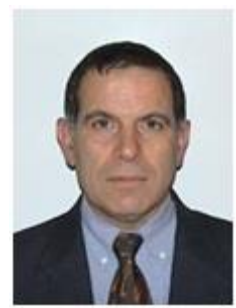

**Ibrahim I. Esat** has BSc and PhD from Queen Mary College of London University, after completing his Phd, he worked on mechanism synthesis at Newcastle university for several years, after this he moved to a newly formed University in Cyprus spending one and half year before returning back to UK joining to the University College of London University where he worked on designing "snake arm" robot. After this he moved to the Dunlop Technology division as a principal engineer working on developing CAD packages in particular surface modellers, following this he returned back to academia as a lecturer at Queen Mary College and later moving to Brunel University. He continued working closely with industry. He continued developing bespoke multi body dynamics software with user base in the UK, Europe and USA. Currently he is a full professor at Brunel University.